\documentclass{article}

\usepackage{mdframed}
\usepackage{microtype}
\usepackage{graphicx}
\usepackage{subfigure}
\usepackage{booktabs} 

\usepackage{hyperref}

\usepackage[accepted]{icml2025}

\usepackage{amsmath}
\usepackage{amssymb}
\usepackage{mathtools}
\usepackage{amsthm}

\usepackage[capitalize,noabbrev]{cleveref}

\theoremstyle{plain}

\theoremstyle{definition}

\theoremstyle{remark}

\usepackage[textsize=tiny]{todonotes}

\icmltitlerunning{Partially Rewriting a Transformer in Natural Language}

\begin{document}

\twocolumn[
\icmltitle{Partially Rewriting a Transformer in Natural Language}



\icmlsetsymbol{equal}{*}

\begin{icmlauthorlist}
\icmlauthor{Gonçalo Paulo}{eai}
\icmlauthor{Nora Belrose}{eai}
\end{icmlauthorlist}

\icmlaffiliation{eai}{EleutherAI}

\icmlcorrespondingauthor{Gonçalo Paulo}{gonçalo@eleuther.ai}

\icmlkeywords{Machine Learning, ICML}

\vskip 0.3in
]



\printAffiliationsAndNotice{}  

\begin{abstract}
The greatest ambition of mechanistic interpretability is to completely rewrite deep neural networks in a format that is more amenable to human understanding, while preserving their behavior and performance. In this paper, we attempt to partially rewrite a large language model using simple natural language explanations. We first approximate one of the feedforward networks in the LLM with a wider MLP with sparsely activating neurons - a transcoder - and use an automated interpretability pipeline to generate explanations for these neurons. We then replace the first layer of this sparse MLP with an LLM-based simulator, which predicts the activation of each neuron given its explanation and the surrounding context. Finally, we measure the degree to which these modifications distort the model's final output. With our pipeline, the model's increase in loss is statistically similar to entirely replacing the sparse MLP output with the zero vector. We employ the same protocol, this time using a sparse autoencoder, on the residual stream of the same layer and obtain similar results. These results suggest that more detailed explanations are needed to improve performance substantially above the zero ablation baseline.
\end{abstract}

\section{Introduction}

While large language models (LLMs) have reached human level performance in many areas \citep{guo2025deepseek}, we understand little about their internal representations. Early mechanistic interpretability research attempted to explain the activation patterns of individual neurons \citep{Olah2020,gurnee2023finding,gurnee2024universal}, but research has found that most neurons are ``polysemantic'', activating in semantically diverse contexts \citep{arora2018linear,elhage2022toy}.

Sparse autoencoders (SAEs) were proposed to address polysemanticity \citep{cunningham2023sparse}. SAEs consist of two parts: an encoder that transforms activation vectors into a sparse, higher-dimensional latent space, and a decoder that projects the latents back into the original space. Both parts are trained jointly to minimize reconstruction error. Recently, a significant effort was made to scale SAE training to larger models, like GPT-4 \citep{gao2024scaling} and Claude 3 Sonnet \citep{templeton2024scaling}, and they have become an important interpretability tool for LLMs.
\citet{paulo2024automatically} took inspiration on \citet{bills2023language} and built an automated pipeline for generating natural language explanations of SAE features and evaluating how good these explanations are, although rigorously measuring how interpretable an explanation is still a complicated and methodologically fraught task.

Recently \citet{dunefsky2024transcoders} proposed sparse \emph{transcoders} as an alternative method for extracting interpretable features from LLMs. The architecture of the transcoder is identical to that of an SAE, but it is trained to predict the output of a feedforward network given its input. We can then entirely replace the original FFN with its transcoder approximation, thereby partially rewriting the model in terms of more interpretable primitives.

\begin{figure*}
    \centering
    \includegraphics[width=1\linewidth]{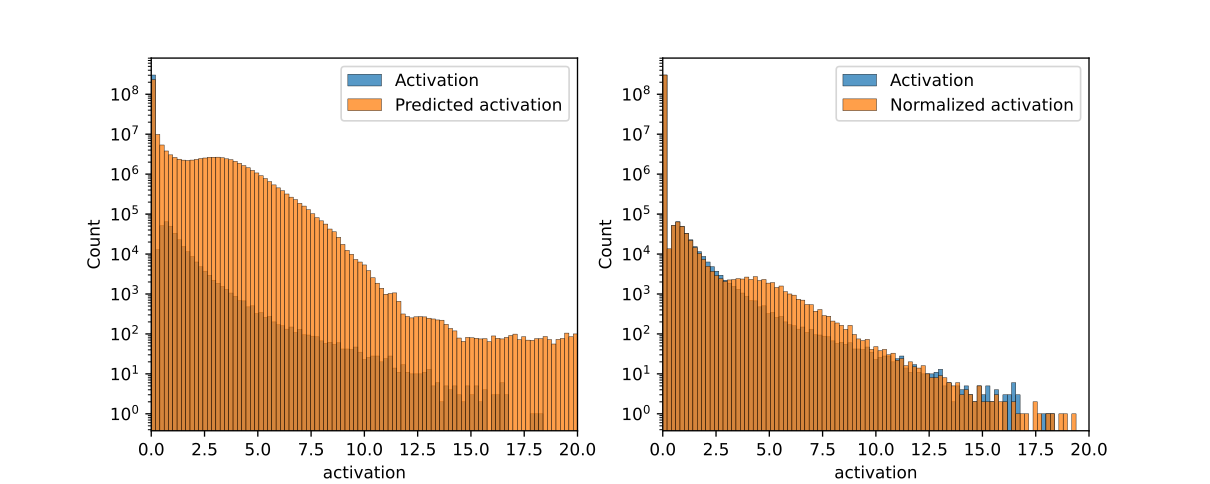}
    \caption{\textbf{Distribution of predicted activations for all latents.} On the left we compare the distribution of predicted activations before normalization, and on the right we show what the distribution looks like after quantile normalization. Before normalization, the predictor model systematically over-predicts high activation values by multiple orders of magnitude. Quantile normalization primarily has the effect of enforcing a prior in favor of features not being active.}
    \label{fig:distribution}
\end{figure*}

The idea of rewriting a neural net in a more interpretable form is not new. The ``microscope AI'' framework \citep{hubinger2019chris} aims to analyze a neural network's learned representations to gain actionable insights for humans, rather than using the network directly. These insights would likely take the form of natural language explanations of the network's features and circuits. Microscope AI aims to reduce risks associated with model deployment while still benefiting from the model's knowledge. Imitative generalization is a proposal to extend this idea by jointly optimizing the network and its human-interpretable annotations to maximize their prior likelihood \citep{barnes2021imitative}.

In this work, we pursue the following idea: if the latents of a transcoder are interpretable enough, we can \emph{simulate} their activations using natural language explanations. Specifically, we replace the encoder of the transcoder with an LLM prompted to predict the activation of each latent given its explanation and the textual context. We then patch this modified transcoder back into the model, hopefully yielding behavior nearly identical to the unpatched model. In the limit, we could use this to ``rewrite'' every feedforward layer in the model in terms of interpretable features and operations on those features.

With our pipeline, the model's increase in loss is only slightly smaller than when entirely replacing the sparse MLP output with the zero vector. Our results suggest that more detailed explanations are needed to improve performance substantially above the zero ablation baseline. There are many potential ways to improve the quality and specificity of explanations, and we hope this work inspires the community to invest more resources in this direction.

\section{Methods}

We begin by training a sparse transcoder on the MLP of the sixth layer of Pythia 160M \citep{biderman2023pythia}. Our loss function is the mean squared error between the transcoder's output and the MLP output, with no auxiliary loss terms. Sparsity is continously enforced on the transcoder latents using the TopK activation function proposed by \citet{gao2024scaling} with $k = 32$. We train over the first 8B tokens of Pythia's training corpus, the Pile \citep{gao2020pile}, using the Adam optimizer \citep{kingma2014adam}, a sequence length of 2049, and a batch size of 64 sequences. We have also added a linear ``skip connection'' to the transcoder, which we find improves its ability to approximate the original MLP at no cost to interpretability scores. That is, the transcoder takes the functional form
\begin{equation}
    f(x) = \mathbf{W}_2 \mathrm{TopK}(\mathbf{W}_1 x + \mathbf{b}_1) + \mathbf{W}_{\mathrm{skip}} x + \mathbf{b}_2
\end{equation}
Both $\mathbf{W}_2$ and $\mathbf{W}_{\mathrm{skip}}$ are zero-initialized, and $\mathbf{b}_2$ is initialized to the empirical mean of the MLP outputs, so that the transcoder is a constant function at the beginning of training. We leave a deeper analysis of the skip connection for future work. We also train a sparse autoencoder on the residual stream, with the same training conditions as the transcoder, but without a skip connection.

\begin{figure*}[t]
    \centering
    \includegraphics[width=1\linewidth]{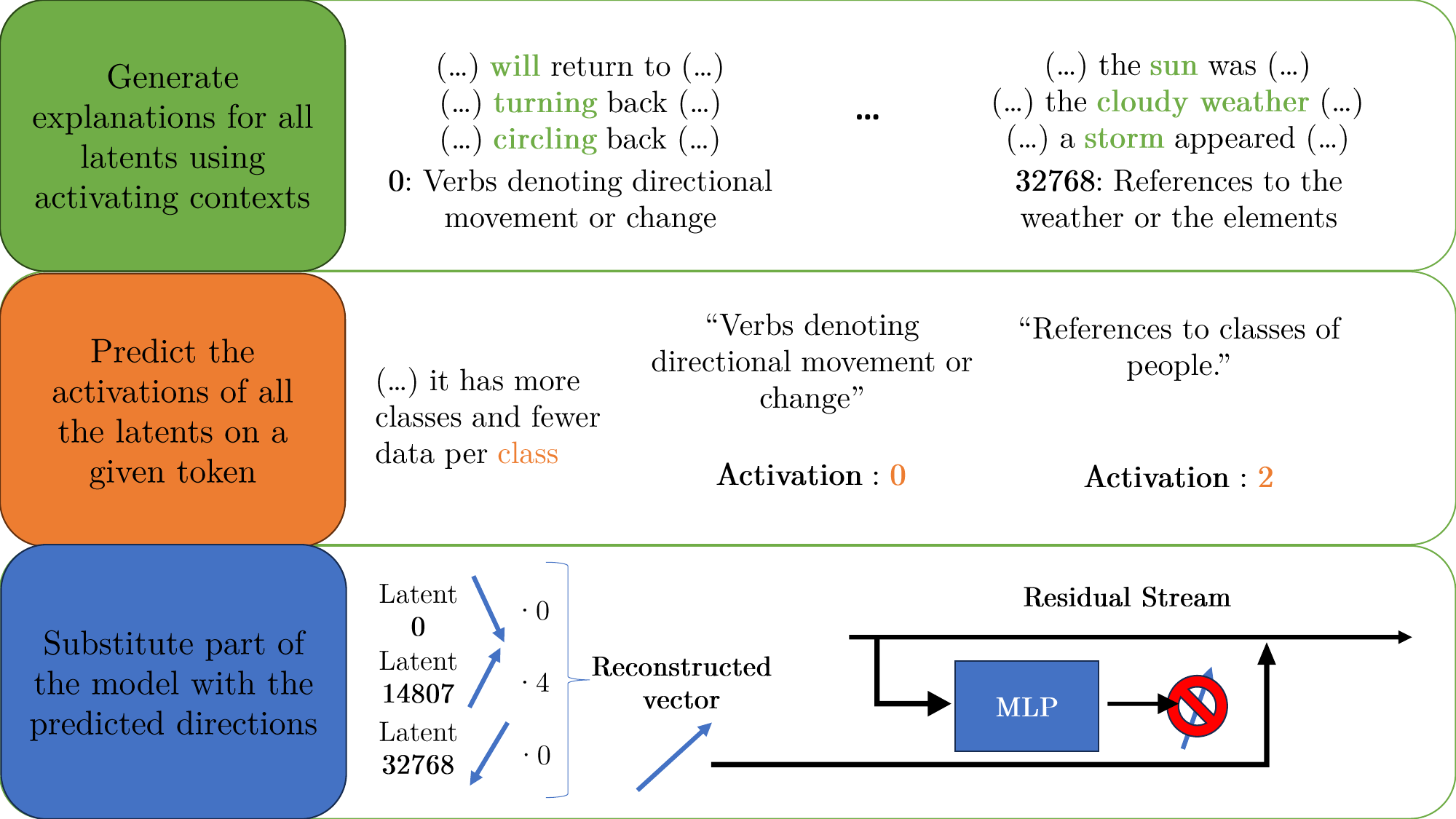}
    \caption{\textbf{Partially rewriting an LLM.} After training a Transcoder, or any type of SAE, we generate explanations for all the latents using the contexts where that latent is active. An LLM is tasked to summarize or otherwise find patterns in the activations and output a simple, single sentence explanation for that latent. These explanations are used by another instance of an LLM to predict wether the latent should be active in a given token. After some post-processing of those predictions, a reconstruction vector is calculated using the decoder directions of the latents that are considered to be active for that token.}
    \label{fig:pipeline_explanation}
\end{figure*}

We use the automated interpretability pipeline released by \citet{paulo2024automatically} to generate explanations and scores for transcoder and SAE latents. Then we modify the pipeline to do ``single'' token simulation, tasking an LLM to determine if a latent is active on the last token of a sequence, and by how much (Figure \ref{fig:pipeline_explanation}).

\subsection{Quantile normalization}\label{sec:quantiles}

We found in early experiments that Llama produces highly uncalibrated predictions of feature activations: the marginal distribution of the predicted activations differs markedly from the marginal distribution of the true activations (\cref{fig:distribution}). Patching these uncalibrated activations into the model yields very poor results. To alleviate this problem, we use \href{https://en.wikipedia.org/wiki/Quantile_normalization}{quantile normalization}, which monotonically transforms the model's predictions in such a way that their marginal distribution matches that of the true activations. This transformation is an optimal transport map under a variety of cost functions \citep{santambrogio2015optimal}. 

We compute the quantile normalizer separately for each individual feature, using the empirical CDFs of the simulator's predicted activations and of the true activations. Since it is vastly more efficient to compute true activations directly from the transcoder, we compute the CDF for these on a much larger dataset of 10M tokens, while we are only able to use 10K contexts (with each context contributing a single token) for the predicted activations. Once the quantile normalizer has been computed, this transformation is then applied to all simulator predictions. We find that the quality of this transformation is strongly dependent on the sample size used for the predicted activations, and the loss of the partially rewritten model is sensitive to this distribution, see discussion in Appendix \ref{app:sample_size}.

One problem with estimating the quantile normalizer on an empirical sample is that, while the empirical CDF is an unbiased estimator for the true CDF, the empirical inverse CDF (or quantile function) is biased for the true inverse CDF. This means that a quantile normalizer fit on a modestly sized dataset of predicted and true activations may generalize poorly to unseen inputs. In a future draft of this paper we plan to experiment with bias-corrected estimators for the population quantiles \citep{hyndman1996sample}, which will hopefully improve the sample complexity. On the other hand, since we have three orders of magnitude more datapoints for the true activation quantiles than we do for the predicted activation CDF, the biased quantiles may matter less than the finite sample variance in the predicted activation statistics.

\begin{figure*}[t]
    \centering
    \includegraphics[width=1\linewidth]{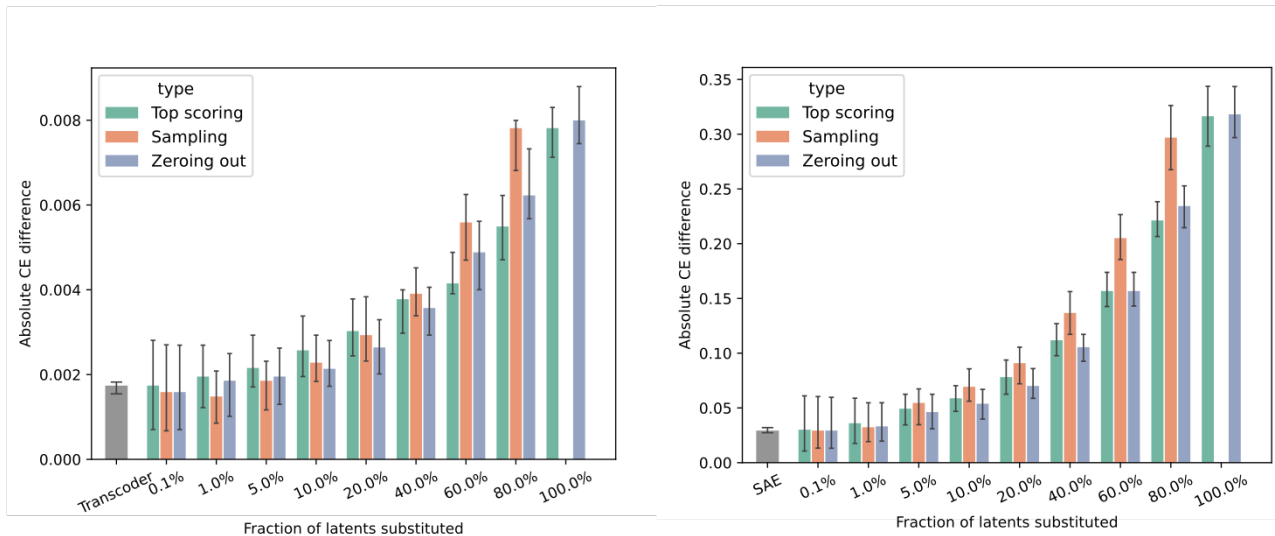}
    \caption{\textbf{Cross entropy loss increase for different fractions of transcoder and SAE substitution.} We compute the CE loss over 10K prompts, for the transcoder (left) and SAE (right) respectively, by substituting parts of the encoder with natural language explanations. Bars in green show the average loss increase when choosing the top scoring latents for replacement. Bars in orange show the average loss increase when randomly selecting a subset of latents to replace. Bars in blue show the average loss increase caused by zeroing out a part of the transcoder. Bar heights represent the median value of the absolute difference, because the distribution is heavy-tailed, and error bars are $95\%$ confidence intervals computed using bootstrapping. The interpretability score used for the selecting latents is detection scoring, \citep[page 5]{paulo2024automatically}, computed over 100 positive and 100 negative samples. Over this set of prompts, Pythia had a cross entropy loss of $3.19 \pm 0.09$ nats per token.  }
    \label{fig:substitution}
\end{figure*}

\section{Evaluation}

Simply replacing a single MLP with a transcoder increases the model's cross-entropy loss to that of an early Pythia checkpoint- namely one that was trained on only 25\% of the data.\footnote{The cross-entropy loss of Pythia 160M checkpoints on this set of prompts is not monotonic with training time, so a more precise estimate is not possible.} Rewriting any part of the transcoder in natural language will necessarily degrade the model's performance even further. Consequently, we focus on rewriting a single MLP block of Pythia 160M, since rewriting all MLP blocks simultaneously would likely cause the model to become completely unusable. The same applies to the residual stream SAE: the performance of the model when adding a single SAE to the residual stream is close to using a checkpoint trained on only 10\% of the original data.

For evaluation, we sample chunks of text from the Pile and gather latents from the transcoder evaluated on the last token of each chunk. For each latent in each text chunk, we prompt Llama 3 Instruct 8B \citep{dubey2024llama} to output a number from zero to ten indicating how strongly it predicts the latent should activate given its natural language explanation and the textual context. We record the probability that Llama assigns to each of the ten numbers, and compute the expected value. This expected prediction is then quantile normalized (\cref{sec:quantiles}) to produce the predicted activation for this latent.

This yields a vector of predicted activation values for each latent, and we apply the TopK activation function to this vector to ensure it has the expected level of sparsity. We evaluate over 10K different prompts for the transcoder and 1K different prompts for the SAE, measuring the cross-entropy loss for next-token prediction.

\paragraph{Partial rewriting.} We also experiment with mixing predicted and ground truth latent activations in varying proportions, allowing us to examine the effect of rewriting only part of the encoder. We do this in two different ways:
\begin{enumerate}
    \item Select the top $k$ most interpretable features according to our evaluation pipeline \citep{paulo2024automatically}. This is labeled ``Top scoring'' in Figure~\ref{fig:substitution}.
    \item Sample $k$ features uniformly at random from the transcoder. This is labeled ``Sampling'' in Figure~\ref{fig:substitution}.
\end{enumerate}

\begin{figure*}[!h]
    \centering
    \includegraphics[width=1\linewidth]{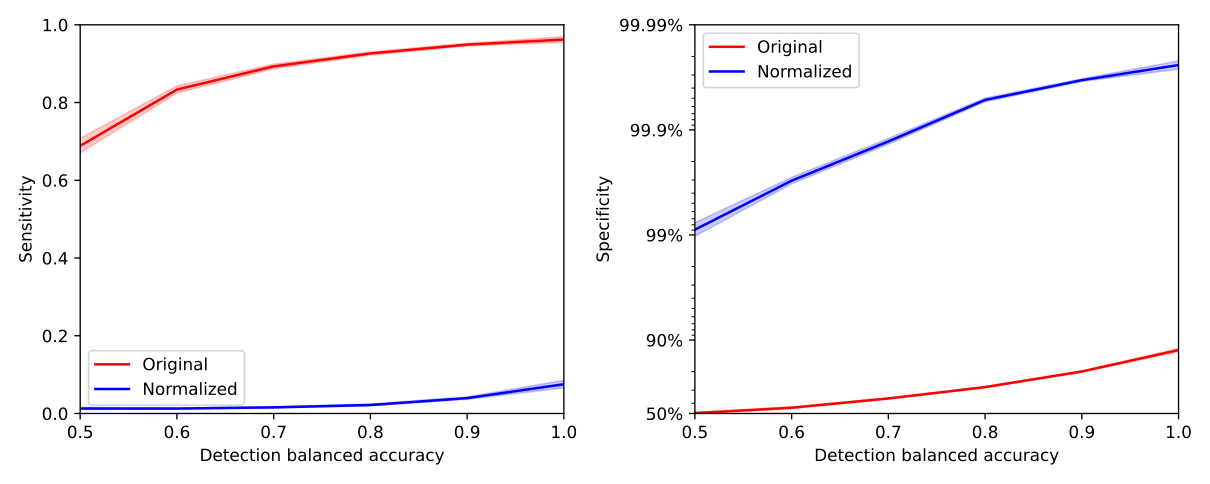}
    \vskip -0.3cm 
    \caption{\textbf{Detection score predicts sensitivity and specificity.} Binning explanations by their scores makes it evident that high-scoring explanations are more specific and sensitive.}
    \label{fig:scores}
\end{figure*}
\section{Results}
\label{sec:results}

Figure \ref{fig:substitution} illustrates how cross-entropy loss increases as we replace more and more transcoder latents with their simulated counterparts. When we replace all latents with simulated counterparts, the cross-entropy loss is the same as that of a Pythia checkpoint trained on only 10-15\% of the full training corpus. This result is similar to that of setting the output of the MLP to the zero vector. This means that a \textbf{predictor that ignores the explanations and always predicts that every latent is inactive would achieve only slightly more loss than this setup}, for any fraction of re-written transcoder and SAE. Randomly selecting which latents to substitute, instead of always substituting the best scoring latents, leads to a bigger performance hit than zeroing out the MLP. Not calibrating the predictions using quantile normalization leads to an even worse performance, equivalent to barely training the model at all (Figure~\ref{fig:not_normalized}). 

Using the empirical distribution as the target distribution for quantile normalization significantly improves this performance and rewritten models perform better than zeroing them, see Figure. Simulating activations over 10K prompts requires individually prompting a model for 32768 latents, for a total of 327 million predictions, an expensive endeavor. We expect that, if performed on 100 thousand, the normalized distribution would match the empirical distribution, see discussion in \cref{app:sample_size}, and that the performance overwriting the model would be better even when using a larger calibration distribution and not the real distribution, but due to the computational costs we cannot claim that for sure.

\subsection{Explanations are not detailed enough}

The poor performance of the model when using the uncalibrated predicted activations is mainly due to the low specificity of explanations. To see this, let's consider the case that the classifier only achieves a specificity of 99\%. The transcoder used in this work has 32768 latents, and if the LLM predictor only can only achieve a specificity of 99\% that means, that on average, it predicts there are 320 active latents, which is 10 times more than the actual number ($k = 32$). Even with a specificity of 100\%, where the model only predicts 32 latents to be active, it is unlikely that the top 32 predictions would be the correct ones. We observe than on average the current automatic latent explanation setup has a specificity of around 80\%, a number much lower than what would be required to do this task. 

By performing quantile normalization, a large chunk of incorrectly predicted activations, activations that should be zero but were given a non-zero value by the predictor, are set back to zero. This significantly increases the specificity, enough that some of the original model's performance is maintained, but at the same time this significantly decreases the sensitivity, as some of the correctly classified active latents are also set to zero. This makes it clear that the current pipeline is lacking, as it is not specific enough when its simulated activations are uncalibrated and is not sensitive enough when they are.

We find that detection scores \citep[page 5]{paulo2024automatically} are predictive of the specificity and sensitivity of an explanation, with higher scoring latents corresponding to explanations that have higher specificity and sensitivity (\cref{fig:scores}). This is expected, as detection scoring corresponds to detecting whether a given latent is active on a given context, which is similar to our simulation task in this work. For the same reason, latents with higher fuzzing scores also have higher sensitivity and specificity (\cref{fig:fuzz_scores}).

\section{Conclusion}

In this work, we proposed a new methodology for rigorously evaluating the faithfulness of natural language explanations of sparse autoencoder transcoder latents, based on partially rewriting the base model using these explanations. We found that existing explanations are severely wanting: without quantile normalization they are insufficiently precise to enable even $20\%$ of latents to be simulated while preserving the base model's performance. While normalization improves performance significantly, we are still unable to outperform the zero ablation baseline, where the entire MLP is replaced with the zero vector.

This is mainly due to the fact that explanations are not specific enough, leading to a high number of false positives. Our results highlight the fact that it is important for an explanation to correctly identify the contexts where a feature is not active, in addition to the feature's activation level in contexts where it is active. Future work on the interpretability of latents should take this into consideration.

To improve upon these results, new techniques are needed to make explanations more specific, for instance using contrast pairs of highly similar features to bring out additional details. This could potentially increase the sensitivity as well, which takes a big hit when using quantile normalization.

\section{Contributions and Acknowledgments}
Nora and Gonçalo had the idea for the simulation experiments. Gonçalo executed the experiments and wrote the first draft of the article. Nora trained the transcoders, suggested the quantile normalization experiments, provided feedback and did significant revisions.

We thank Alice Rigg for discussions and comments. Gonçalo and Nora are funded by a \href{https://www.openphilanthropy.org/grants/eleuther-ai-interpretability-research/}{grant} from Open Philanthropy. We thank Coreweave for computing resources.

\section{Code Availability}

Code for these experiments is available on the \texttt{nl\_simulations} branch of the \href{https://github.com/EleutherAI/sae-auto-interp}{sae-auto-interp} GitHub repo.

\section*{Impact Statement}

This paper presents work whose goal is to advance the field of Mechanistic Interpretability. There are many potential societal consequences 
of our work, none which we feel must be specifically highlighted here.

\bibliography{bibliography}
\bibliographystyle{icml2025}

\clearpage
\appendix
\renewcommand{\thefigure}{A\arabic{figure}}
\setcounter{figure}{0}

\section{Simulation prompt}
\begin{mdframed}
\begin{verbatim}
You are an intelligent and
meticulous linguistics researcher.

You will be given a  certain
explanation of a feature of
text, such as "male pronouns" 
or "text with negative sentiment" 
and examples of text that contains
this feature. Some explanations
will be given a score from 0 to 1.
The higher the score the better
the explanation is, and you 
should be more certain 
of your response (positive 
or negative).

These features of text are normally
identified by looking for specific
words or patterns in the text.
There are many features associated 
with a single token, and sometimes 
the feature is related with 
the previous token or context.

Your job is to identify how much 
the last token,  which is marked 
between << and >>, represents 
the feature.  You will output 
a integer between 0 and 9, 
where 0 corresponds to no relation 
to the explanation and 9 to a 
strong relation.

Most of the tokens should have
no relation. The ones that 
are related, should more likely
be given 1 than 2, 2 than 3, 
and so on. Only give a 9 if the 
description exactly matches the token.

You must return your 
response in a valid Python list.
Do not return anything
else besides a Python list.

\end{verbatim}
\end{mdframed}

\section{Sample size}
\label{app:sample_size}

\begin{figure*}[t]
    \centering
    \includegraphics[width=1\linewidth]{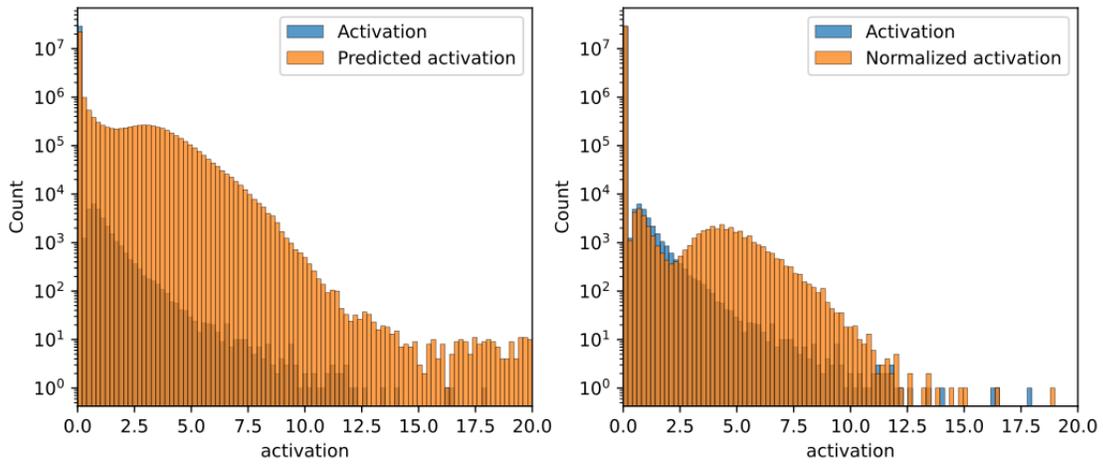}
    \caption{\textbf{Distribution of predicted activations for all latents over a smaller sample size} If we use only 1K prompts as the predicted activations and use the 10M prompts as the target distribution, the mismatch with the empirical activation distribution is higher.}
    \label{fig:sample_size}
\end{figure*}

Activations of latents are very infrequent, and the empirical distribution of their activations on a small number of tokens can have a small number of non-zero values. If we use only 1K samples instead of 10K samples, the mismatch between the normalized and the empirical distribution \ref{fig:distribution} grows larger (\cref{fig:sample_size}). We then expect that a larger number of samples would lead to a better convergence. 

\begin{figure}[t]
    \centering
    \includegraphics[width=1\linewidth]{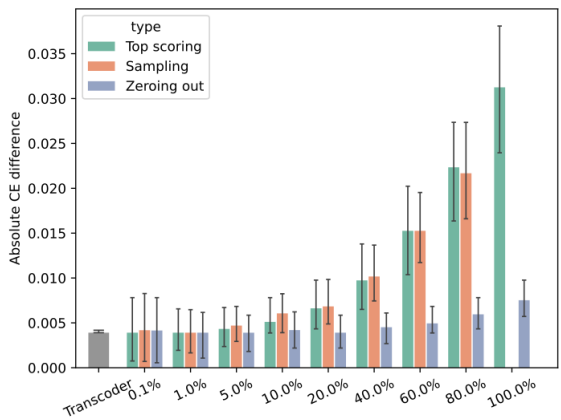}
    \caption{\textbf{Cross entropy loss increase for different fractions of transcode depends on the sample size}. Using only 1K samples to compute the quantile normalization function leads to a much worse CE loss when performing substitution, due to the larger difference between the empirical and the normalized distribution (\cref{app:sample_size}) }
    \label{fig:predict_figure}
\end{figure}

We observe that the fact that the normalized activation distribution and the empirical activation distribution not matching significantly worsens the performance of substituting the predicted activations as the performance over the 1K samples, using the distribution in \cref{fig:sample_size}, is much worse ( \cref{fig:predict_figure}).

\begin{figure*}[h]
    \centering
    \includegraphics[width=1\linewidth]{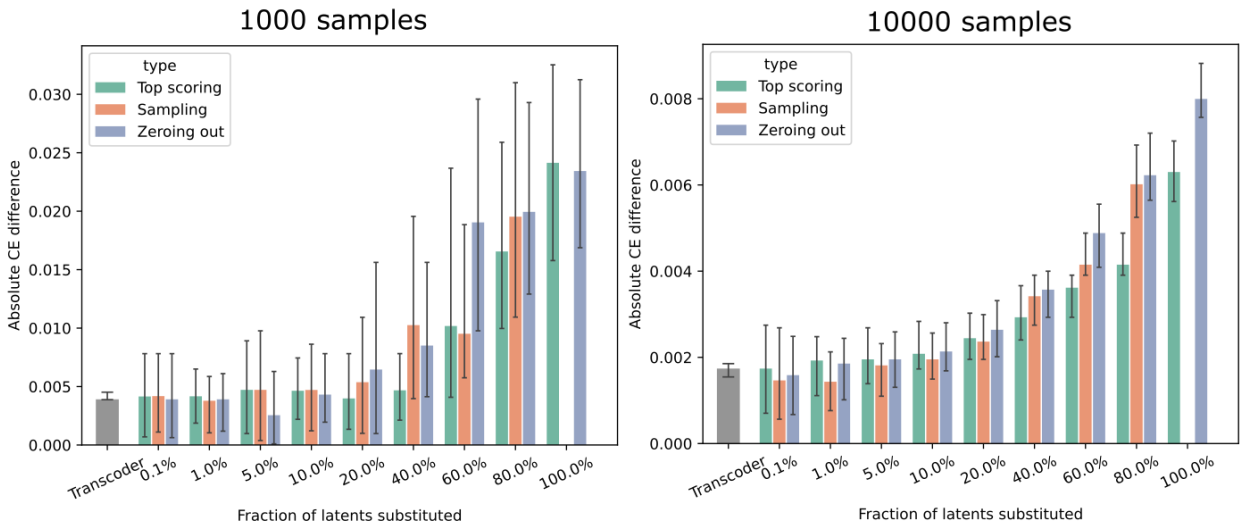}
    \caption{\textbf{Using the empirical distribution over the prompt improves results.} If instead of using the empirical distribution of activations computed over the larger 10M token sample, we use the empirical distribution of activations that we are predicting, the results of substituting the predictions improve significantly. We expect that, as the number of samples increases, these should converge, and that substituting predictions will perform better than zeroing out the component.}
    \label{fig:predict_cheating}
\end{figure*}

Using the empirical activation distribution over the prompts improves the result both in the 1K samples case and in the 10K samples case (\cref{fig:predict_cheating}). We expect that, were we able to have done prediction over 100k samples, the results of substituting using the empirical distribution over the prompts or over the larger 10M token sample would be closer, and if the trend holds, substituting the predictions would be better than zeroing out the MLP.

\begin{figure*}
    \centering
    \includegraphics[width=1\linewidth]{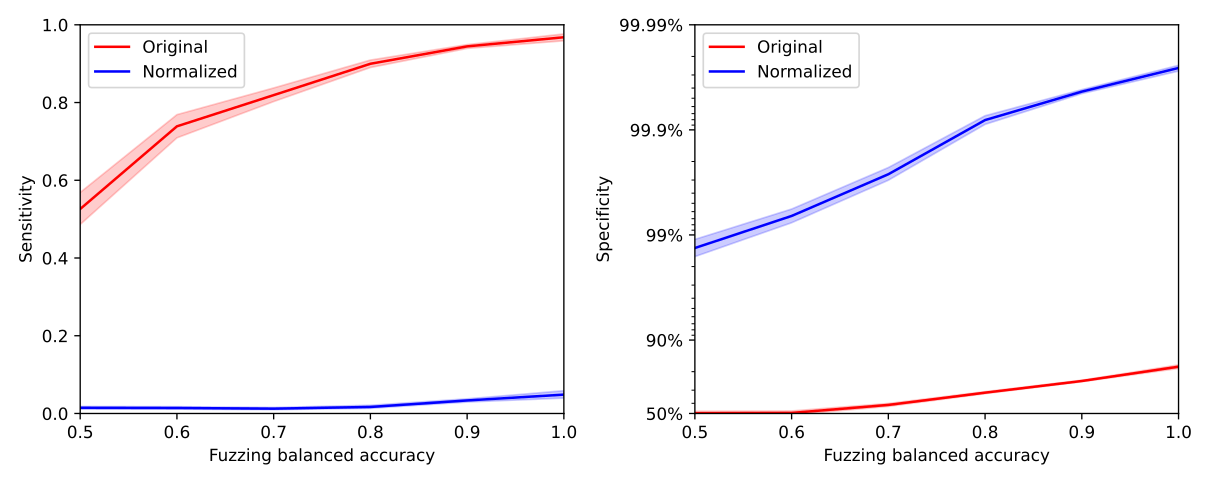}
    \caption{\textbf{Fuzzing score predicts sensitivity and specificity} Explanations with higher fuzzing scores lead to better predictions of the simulations}
    \label{fig:fuzz_scores}
\end{figure*}

\begin{figure}[t]
    \centering
    \includegraphics[width=1\linewidth]{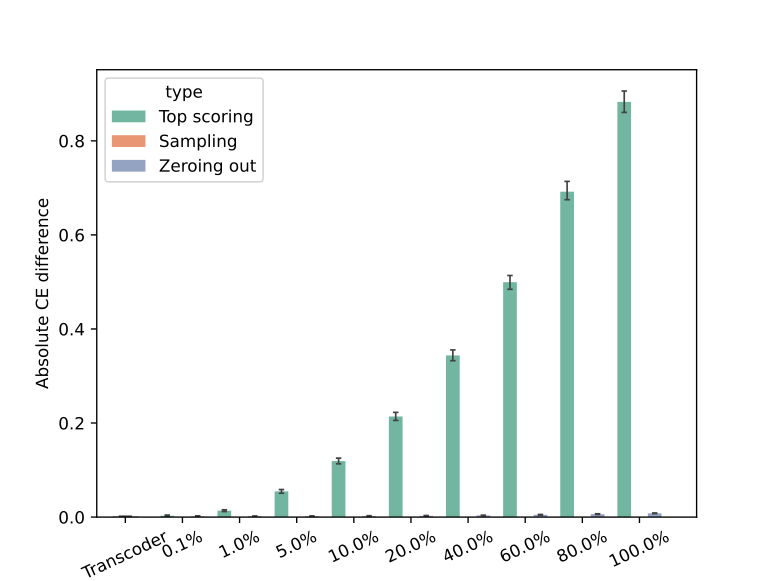}
    \caption{\textbf{Cross entropy loss increase for different fractions of transcoder substitution.} Not normalizing the predicted activations leads to a much worse CE loss than normalizing (\cref{fig:substitution}). The interpretability score used for the selecting latents is detection scoring, \citep[page 5]{paulo2024automatically}, computed over 100 positive and 100 negative samples. Over this set of 1K prompts, Pythia had a cross entropy loss of $3.19 \pm 0.09$. Horizontal lines correspond to the CE difference of different Pythia checkpoints trained on less of data. }
    \label{fig:not_normalized}
\end{figure}

\end{document}